\documentclass[final]{cvpr}

\usepackage{times}
\usepackage{epsfig}
\usepackage{graphicx}
\usepackage{amsmath}
\usepackage{amssymb}

\usepackage{enumitem}
\setlist[itemize]{leftmargin=5.5mm}
\usepackage{amsmath}
\usepackage{multicol}
\usepackage{multirow}
\usepackage{afterpage}
\usepackage{url}            
\usepackage{booktabs}       
\usepackage{amsfonts}       
\usepackage{nicefrac}       

\usepackage[pagebackref=true,breaklinks=true,colorlinks,bookmarks=false]{hyperref}

\def\cvprPaperID{****} 
\def\confYear{CVPR 2021}

\begin{document}
\graphicspath{{figures/}}
\title{Halluci-Net: Scene Completion by Exploiting Object Co-occurrence Relationships}
\author{
    Kuldeep Kulkarni$^1$\thanks{This work was done while Kuldeep and Tejas were at CMU.} \quad
    Tejas Gokhale$^2$\footnotemark[1] \quad 
    Rajhans Singh$^2$ \quad 
    Pavan Turaga$^2$ \quad 
    Aswin Sankaranarayanan$^3$
\\$^1$Adobe Research India, $^2$Arizona State University, $^3$Carnegie Mellon University
\\{
    \tt\small kulkulka@adobe.com, 
    \{tgokhale, rsingh70, pturaga\}@asu.edu, 
    saswin@andrew.cmu.edu
}
}
\maketitle

\begin{abstract}
Recently, there has been substantial progress in image synthesis from semantic labelmaps.
However, methods used for this task assume the availability of complete and unambiguous labelmaps, with instance boundaries of objects, and class labels for each pixel.
This reliance on heavily annotated inputs restricts the application of image synthesis techniques to real-world applications, especially under uncertainty due to weather, occlusion, or noise. 
On the other hand, algorithms that can synthesize images from sparse labelmaps or sketches are highly desirable as tools that can guide content creators and artists to quickly generate scenes by simply specifying locations of a few objects.
In this paper, we address the problem of complex scene completion from sparse labelmaps.
Under this setting, very few details about the scene (30\% of object instances) are available as input for image synthesis. 
We propose a two-stage deep network based method, called `Halluci-Net', that learns co-occurence relationships between objects in scenes, and then exploits these relationships to produce a dense and complete labelmap.
The generated dense labelmap can then be used as input by state-of-the-art image synthesis techniques like pix2pixHD to obtain the final image.
The proposed method is evaluated on the Cityscapes dataset and it outperforms two baselines methods on performance metrics like Fr\'echet Inception Distance (FID), semantic segmentation accuracy, and similarity in object co-occurrences.
We also show qualitative results on a subset of ADE20K dataset that contains bedroom images.
\end{abstract}

\section{Introduction}

\begin{figure}[t]
    \centering
    \includegraphics[width=\linewidth]{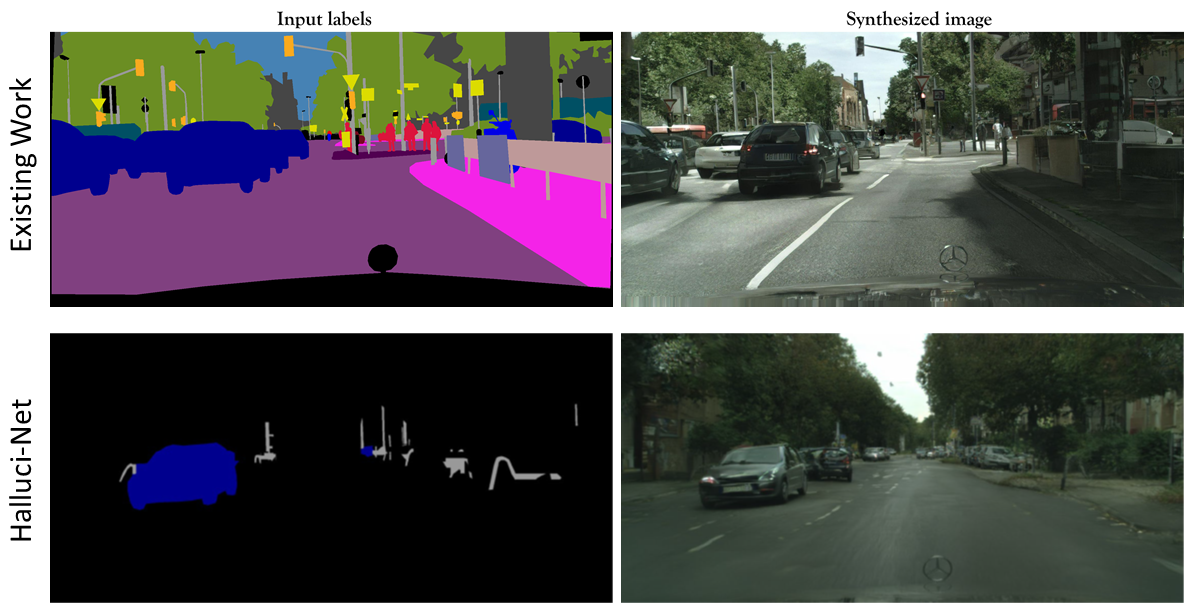}
    \caption{This paper addresses the task of image synthesis from sparse labelmaps, in contrast to existing work which assumes complete labelmaps as inputs.
    This work is a step towards image synthesis models that can hallucinate the complete scene when sparse and incomplete inputs are encountered.}
    \label{fig:motivation}
\end{figure}

\begin{figure*}
    \centering
    \includegraphics[ width=\textwidth]{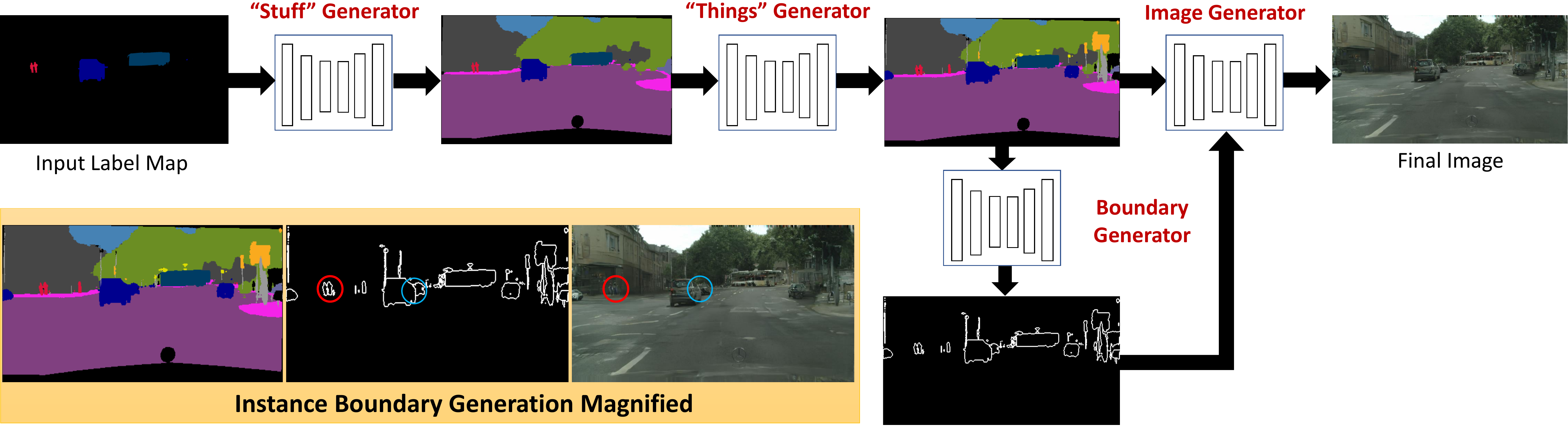}
    \caption{Overview of the algorithm: for both stages, we train a GAN using a modified pix2pix
    based generator.
    Inset shows details of generated instance boundary maps.
    The instance boundaries created between adjacent cars and adjacent persons are in blue and red circles respectively. 
    Boundary maps help create distinguishable instances of the same object class in the synthesized images.}
    \label{fig:overview}
\end{figure*}

Reconstruction of scenes has been a project that humankind has undertaken since pre-historic times.
While democratization of cameras has exponentially improved our ability to capture images, recently there has been a growing body of work on synthesizing images from labelmaps.
Image synthesis techniques, powered by generative models~\cite{goodfellow2014generative}, have led the progress of converting high-level inputs to images, such as for synthesizing images from labels~\cite{isola2017image,wang2018high,park2019semantic} or for converting sketches to images~\cite{chen2018sketchygan,sangkloy2017scribbler}, by designing an image-to-image translation framework using conditional GANs.
However, complex scenes like streets and bedrooms are often composed of several objects that require very precise pixel-level information in the form of dense labelmaps or sketches.
While fully-supervised methods such as pix2pixHD~\cite{wang2018high}, GauGAN~\cite{park2019semantic} are able to generate high-quality photorealistic images from dense labelmaps, they do not possess the ability to do so when only partial labelmaps are available.

In many cases, a complete labelmap of a scene is not available, but partial information such as locations of some objects may be available.
Humans (or rather, some artistic humans~\cite{cohen1997can}) are able to imagine and recreate scenes from partial memory.
However artists need to spend prohibitively large amounts of time at creating scenes using conventional art media such as paints, watercolor, ink or charcoal.
It is therefore highly desirable to empower them with the ability to create images of new scenes with desired characteristics and constituent objects, while minimizing the amount of effort needed.
An image synthesis technique (as shown in Figure~\ref{fig:motivation}) that can receive an input with only a few known object locations and classes, and generate the image for the complete scene, is a step towards this goal.

This paper addresses this problem of scene completion from sparse labelmaps or labelled sketches. 
Under this setting, the complete labelmap is not available, but the locations of only some of object instances are available as shown in Figure~\ref{fig:motivation}.
While existing work in image synthesis from sparse sketches~\cite{chen2018sketchygan,ghosh2019interactive} only deals with single objects such as shoes or handbags, our work can handle complex scenes with multiple objects and layouts.

To generate the entire scene, one must first learn to speculate about the complete labelmap, i.e.\ the locations and class labels of other object instances.
We divide this problem into several stages as shown in Figure~\ref{fig:overview}.
The first stage takes sparse labelmaps as input, and generates a labelmap for the so-called ``stuffs'' classes (such as road, sidewalk, sky) which provide global scene context.
The second stage then uses both the sparse labelmap and the stuffs labelmap to generate a labelmap for the so-called ``things'' class, that contains fine-grained object classes such as pedestrians, cars, and traffic signs.
We then generate instance boundaries for the ``things'' class.
These instance boundaries and the hallucinated labelmap are then utilized for synthesizing a high-resolution RGB image.
Halluci-Net therefore first learns object co-occurence relationships, and exploits them to convert a sparse labelmap to a complete labelmap and instance boundaries, followed by image synthesis.

The contributions of this paper are summarized below:
\begin{itemize}[nosep,noitemsep]
    \item We introduce a new problem of complex scene completion from extremely sparse inputs such as labelmaps or labelled sketches.
    \item We propose a two-stage method, `Halluci-Net' that captures the object co-occurrence relationships to generate dense labelmaps.
    \item We propose a method to create object instance boundary maps from semantic labelmaps, that are utilized to generate images with well-defined boundaries between different instances of the same object class.
    \item Our two-stage method beats the baseline single-stage method by a considerable margin on various metrics like FID, semantic segmentation measures, similarity in object co-occurrence metrics and human preference scores, and the direct labelmap to image (zero-stage) method in terms of FID.
\end{itemize}

\section{Related Work}
\noindent\textbf{Generative Adversarial Networks} (GANs) have been used to generate images using random inputs seeds~\cite{radford2015unsupervised}.
In order to learn the distribution of images, GANs employ two competing networks -- a generator that outputs a synthesized image, and a discriminator that predicts whether an image is from the dataset or is synthesized.
GANs have been used for numerous vision tasks such as image editing~\cite{zhu2016generative}, image inpainting~\cite{yeh2017semantic}, and super-resolution~\cite{ledig2017photo}.\\

\noindent\textbf{Conditional GANs} 
(cGANs)~\cite{mirza2014conditional} have proved to be useful in the task of image-to-image translation, 
such as generating masked regions of an image~\cite{pathak2016context}, high-resolution face images~\cite{karras2017progressive}, and text-to-images synthesis~\cite{zhang2017stackgan}.
pix2pix~\cite{isola2017image} presents cGANs as a generic solution for numerous image-to-image translation tasks, such as generating images from labelmaps, objects from edge maps, image colorization, thermal images to color images, maps to aerial photos, and day-time images to night-time images.
pix2pixHD~\cite{wang2018high} improves upon pix2pix by generating high resolution $(2048 X 1024)$ and visually appealing images.
It incorporates information regarding object instances (in addition to semantic labelmaps) so as to create well-defined boundaries between object instances of the same category. 
However, the pix2pixHD architecture is prone to producing similar results for different object categories if the input labelmap is too homogeneous.
Park~\etal~\cite{park2019semantic} overcome these short-comings by introducing the spatially-adaptive conditional normalization (SPADE).
Hong~\etal~\cite{hong2018learning} propose a method to manipulate an image with addition and deletion of objects, given the 
locations and object categories to be manipulated.
Liu~\etal~\cite{liu2019learning} propose a generator predicts convolutional kernels conditioned on the semantic labelmap, and a 
discriminator that enhances the semantic alignments between generated images and labelmaps, as compared to the multi-scale discriminator used in pix2pixHD.

\noindent{$ $}

\noindent\textbf{Sketch-to-Image Generation:}
SketchyGAN~\cite{chen2018sketchygan} and ContextualGAN address the problem of generating images from hand-drawn sketches.
This problem considered to be harder than generating images from edges because of the pixel-wise misalignment between sketches and images.
However this work deals only with iconic images with single objects (such as faces, birds, cars, shoes, and handbags).
More recently GAN-based sketch-to-image methods have been extended to the more diverse COCO dataset~\cite{Gao_2020_CVPR}.
We seek to solve an even harder problem of image synthesis from sparse labelmaps, 
i.e. when very few pixels are annotated with labels, which requires learning co-occurrence of objects to hallucinate the entire scene.

\section{Object Co-occurrence Networks}

We propose a two-stage approach -- ``Halluci-Net'' to generate dense labelmaps from sparse labelmaps.
\subsection{Baseline Methods}
Before we discuss our two-stage method, we present two baselines for comparison:
zero-stage and single-stage.
\fbox{
    
    \parbox{0.94\linewidth}{
        \centering
        \textbf{Zero Stage}: Sparse map $\rightarrow$ Image\\
        \textbf{Single Stage}: Sparse map $\rightarrow$ Dense map $\rightarrow$ Image
    }
}
    
    \noindent{$ $}
    
    \noindent\textbf{Zero-stage baseline method:}
    The zero-stage method consists of a single generative adversarial network that directly maps the sparse input labelmap of size $C{\times}H{\times}W$ (where $C$ is the number of classes, $H$ and $W$ are the height and width of the labelmap respectively) to image space. The generator architecture is same as pix2pixHD~\cite{wang2018high} and it is trained with loss functions similar to the pix2pixHD. This method yields reasonable quality images, but falls short of producing new objects, especially the `things' classes.
    
    \noindent{$ $}
    
    \noindent\textbf{Single-stage baseline method:}
    To overcome this limitation, we train a GAN with generator $G_{single}$ conditioned on the sparse labelmap of size $C{\times}H{\times}W$, where $C, H, W$ are the number of classes, height and width of the labelmap respectively.
    The input label $s(c,x,y)$ equals $1$, if the pixel at $(x,y)$ belongs to class $c$, otherwise it is zero. 
    The generator architecture is similar to U-Net~\cite{ronneberger2015u}, except for the first layer, wherein the number of input channels is given by the number of classes. 
    The output of the generator is a dense labelmap 
    $g(c,x,y)$ 
    of size $C{\times}H{\times}W$, the same size as the input. 
    We use a sigmoid function as the last layer to enforce $g(c,x,y) \in [ 0,1]$.
    We use a multi-scale discriminator with a least-square adversarial loss, similar to pix2pixHD.
    
    \subsection{Halluci-Net: Two-stage method}
    \noindent First, we divide the class labels into two categories: 
    \begin{enumerate}[nosep,noitemsep]
        \item \textit{stuffs}: 
        background entities such as road, building, tree, sky, \dots
        \item \textit{things}: 
        foreground objects like car, person, bike, \dots
    \end{enumerate}
    Our input sparse labelmap consists only of ``things''.
    
    Our proposed two-stage method consists of two generative adversarial networks.
    The first stage generator -- $G_{1}$, takes a sparse labelmap of size $(C_{\textit{\small things}} + 1) \times H \times W$, and generates the ``stuffs'' labelmap of $C_{\textit{\small stuff}} \times H \times W$, i.e. predicts pixel locations corresponding to various stuff classes.
    We use the U-net architecture~\cite{ronneberger2015u} for $G_{\textit{1}}$ and train it with an adversarial loss conditioned on the sparse labelmap. 
    
    We overlay the output of the first stage network with the input sparse labelmap and feed it to the second stage generator $G_{\textit{2}}$. 
    This stage has an architecture similar to $G_{\textit{1}}$ except for the number of the channels in the input and output.
    The input of this generator is the overlayed labelmap of size $C \times H \times W$. 
    The output of the second stage generator consists of ``things'' class labelmap of size $C_{\textit{things}}+1 \times H \times W$, where $C_{\textit{things}}$ is the number of ``things'' classes. 
    We then overlay the second stage input and the second stage output to get the final dense map of size $C \times H \times W$. 
    The first and the second stage generators are trained independently. 
    As in the single-stage method, we use the multi-scale discriminator with a least-square adversarial loss. 

\noindent{$ $}

\noindent\textbf{Loss functions:}
For both the baseline and two-stage method, we use the object co-occurrence loss, $L_{OC}$ in~(\ref{eq:oc}) to train the generator. 
Given the generated labelmap $g^*$ and the ground-truth labelmap $g$, 
$L_{OC}$ consists of a standard adversarial loss $L_{adv}$, focal loss $L_{FL}$ (Equation~\ref{eq:fl}), 
and a discriminator feature loss $L_{FM}$ similar to pix2pixHD~\cite{wang2018high}.
\begin{equation}
    L_{OC} = L_{adv}(g^*,g) + \lambda_{FL}L_{FL}(g^*,g) + \lambda_{f}L_{FM}(g^*,g) 
    \label{eq:oc}
\end{equation}
In semantic segmentation tasks, cross-entropy loss function is typically used for the labelmaps.
However since it did not achieve desirable results for our task, we use the focal loss:
\begin{align}
    &L_{FL}(g^*,g) = \sum_{x,y,c}l_{FL}(g^*(c,x,y),g(c,x,y)),~\textit{where}\\
    &l_{FL}(p,y) = -y(1{-}p)^{\gamma}\log p - (1{-}y) p^{\gamma}\log (1-p).
    \label{eq:fl}
\end{align}
Focal loss has been shown to help in the detection of hard-to-classify objects in presence of a large number of easy-to-classify objects~\cite{lin2017focal}.
In our case, it helps in generating rare or hard-to-generate classes like ``biker'' and ``vegetation'' in presence of common or easy-to-generate classes like ``road'', ``tree'', ``side-walk'', etc.

\section{Instance Boundary Generation}
While the proposed two-stage approach can generate plausible dense labelmaps, the nature of the output does not allow us to differentiate between different instances of the same object category. 
In other words, a blob of pixels indicated as belonging to a certain object category may actually consist of pixels of multiple instances of the same category. 
This can be seen in Figure~\ref{fig:overview} where, in the labelmap, there are blobs of ``car'' and ``person'' categories large enough to consist of multiple instances. 
Wang~\etal~\cite{wang2018high} show that when labelmaps with such a characteristic are fed into image generators like pix2pixHD, the boundary between two objects of the same class in the image generated is unclear and the objects generated tend to merge into each other. 
Hence, such blobs need to `split' to create multiple instances and the instance information needs to be incorporated to generate well-defined boundaries in the image. 
Wang~\etal~\cite{wang2018high} also propose to use instance boundary maps as a proxy for incorporating instance information to synthesize images with well-defined boundaries between instances of the same object category. 
To produce such instance boundary maps, they use the ground-truth instance map information, where a pixel is assigned as a boundary pixel if at least one of its four neighbors has a different instance ID.

However, in our case, since the inputs to the image generators are ``hallucinated'' labelmaps (generated from sparse labelmaps), we do not have access to instance IDs.
To circumvent this, we train a GAN to map labelmaps to the corresponding instance boundary maps without ever creating the instance IDs. 
The ground truth to train the instance boundary maps is readily available from the instance ID maps in the dataset. 
The generator architecture is the same as U-Net, except for a two-channel output and the first layer that reflects the size of the input label.
To train the generator for the instance boundary map, we use the loss function, $L_{bd}$ similar to Equation~\ref{eq:oc}, given below:
\begin{align}
     L_{\textit{bd}}(b^*,b) = &~~ 
     \lambda_{\textit{FL}} L_{\textit{FL}}(b^*,b) + \lambda_{\textit{FM}} L_{\textit{FM}} (b^*,b)\nonumber\\
     &+L_{\textit{adv}}(b^*,b) +\lambda_{\text{VGG}} L_{\text{VGG}}(b^*,b) \label{eq:bd}
\end{align}
$L_{adv}$ is the standard adversarial loss, $L_{FL}(b^*,b)$ is the focal loss between the generated boundary map, $b^*=G_{bd}(g^*)$ and the ground-truth $b=G_{bd}(g)$.
We also add VGG feature loss, $L_{VGG}$ for the boundary maps by replicating the boundary maps thrice to form a 3-channel input to compute VGG loss, similar to Wang~\etal~\cite{wang2018high}. 

Figure~\ref{fig:overview} shows the labelmaps and the corresponding instance boundary maps generated by our network. 
The large blob corresponding to the car is split by our network into multiple instances by creating boundaries (marked by blue circle), and similarly the blob of ``person'' class is split into multiple instances (marked by a red circle).

\section{Experiments}
We evaluate our approach to scene completion on the two datasets: Cityscapes~\cite{cordts2016Cityscapes} and ADE20K~\cite{zhou2017scene}. 
The Cityscapes dataset consists of $2975$ examples in the training set and $500$ examples in the validation set. The object classes in this dataset, as mentioned earlier can be divided into two categories, ``things'' like cars, biker, person etc., and ``stuff'' like trees, road, side-walk etc. 
In ADE20K dataset, we consider only bedroom scene images for our experiments.
This bedroom scene set consists of $1389$ examples for training and $139$ examples for validation. 
The ADE20K dataset consists of very large number of object classes but we consider only top-30 object classes (based on the frequency of occurrence per image) and we group the rest into a single class. 
Similar to Cityscapes dataset, we also divide the ADE20K dataset object classes into two categories, ``things'' like bed, table, lamp etc., and ``stuff'' like wall, floor, ceiling etc.
However, these datasets provide dense labelmap and image pairs, instead of sparse labelmaps inputs required for our experiments. To this end, we create our own train/validation datasets to evaluate our approach to scene completion. 

\subsection{Dataset Preparation}

\begin{figure}[t]
     \centering
     \includegraphics[width=\linewidth]{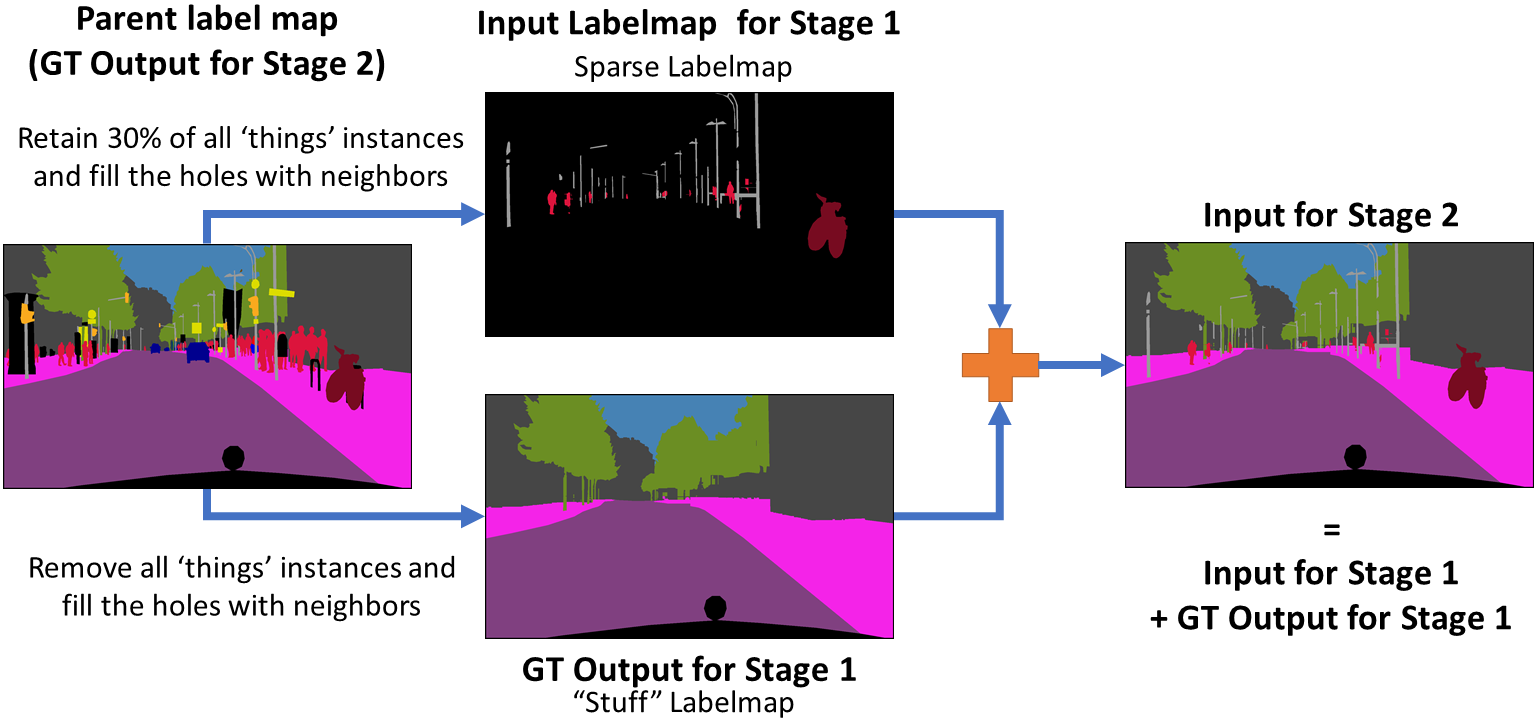}
     \caption{This dataset creation pipeline shows the creation of input-output pairs from existing datasets for training Halluci-Net.
     }
    \label{fig:dataset_creation}
\end{figure}

There is no dataset that is readily available to train the two stages of `Halluci-Net'. 
We first need sufficiently large number of \textit{(sparse labelmap, dense labelmap)} pairs. 
The Cityscapes and the ADE20K datasets were primarily released for the task of semantic segmentation, and contain
pixel-wise labelmaps, instance ID maps and the corresponding images.
We observe the average number of object instances per training example across the dataset is ${\sim}32$ in Cityscapes and ${\sim}31$ in ADE20K. 
Given a parent dense labelmap in the original dataset, we first remove all pixels belonging to ``stuff'' categories and retain $30\%$ random object instances from ``things'' categories. 
Thus, it is possible to create a large number of sparse labelmaps, for every parent dense labelmap, by randomly sampling $30\%$ ``things'' in each image.
Each sparse labelmap and corresponding dense labelmap forms the expected input-output pair for our approach. 
In Figure~\ref{fig:dataset_creation}, we illustrate our data creation pipeline.

\begin{table}
    \centering
    \small
    \begin{tabular}{c ccc}
        \toprule
        \textbf{Method}  & Zero-Stage & Single-Stage & Halluci-Net\\
        \toprule
        w/o inst-bd  & 46.21 & 44.74 & 41.19 \\
        with inst-bd & -     & 42.05 & \textbf{40.67}\\
        \bottomrule
    \end{tabular}
    \caption{Comparison of FID scores (lower $\Rightarrow$ better) with and without instance boundary on Cityscapes~\cite{cordts2016Cityscapes}. 
    Halluci-Net with inst-bd is the best, while 
    Halluci-Net w/o inst-bd also performs considerably better than the single stage approaches.}
    \label{table:fid_vs_fsd}
\end{table}
\begin{table*}[t]
    \centering
    \small
    \begin{tabular}{@{}lccccccc@{}}
        \toprule
        Method   &  \hphantom & \small{(Car, Parking)} & \small{(Person, Person)} & \small{(Pole, Building)} & \small{(Bicycle, Rider)} & \small{(Person, Sidewalk)} & \small{(Car, Bus)}\\
        \midrule
        Halluci-Net     &&   \textbf{0.97}  & \textbf{0.93} &  \textbf{0.89} & \textbf{0.71} & 0.95 & \textbf{0.93}\\
        Single Stage    &&    0.84  & 0.92 & 0.54 & 0.53 & \textbf{0.96} & 0.80\\  
        \bottomrule
    \end{tabular}
    \caption{The table shows similarity between the training and generated sets in terms of object co-occurrences for various pairs of input and output object classes on the Cityscapes~\cite{cordts2016Cityscapes}}
    \label{table:obj_co}
\end{table*}
\subsection{Training}
We train all the networks for 200 epochs. 
In each epoch, for a parent dense labelmap in the original dataset, we use a 
different set of randomly chosen $30\%$ ``things'' categories, and remove all ``stuff'' classes. 
We fix the resolution to $512 \times 256$ for Cityscapes and $512 \times 384$ for ADE20K.
We use Adam optimizer~\cite{kingma2014adam} with $\beta_{1} = 0.5$, $\beta_{2} = 0.999$, learning rate of 0.001 with linear decay after 100 epochs. 
$\lambda_{f}$ and $\lambda_{VGG}$ (if used) are set to 10, and $\lambda_{fl}, \gamma$ to 5 for all experiments.
The instance boundary network is also trained with the same set of hyper-parameters.

\begin{figure}[t]
    \centering
    \includegraphics[width=0.94\linewidth]{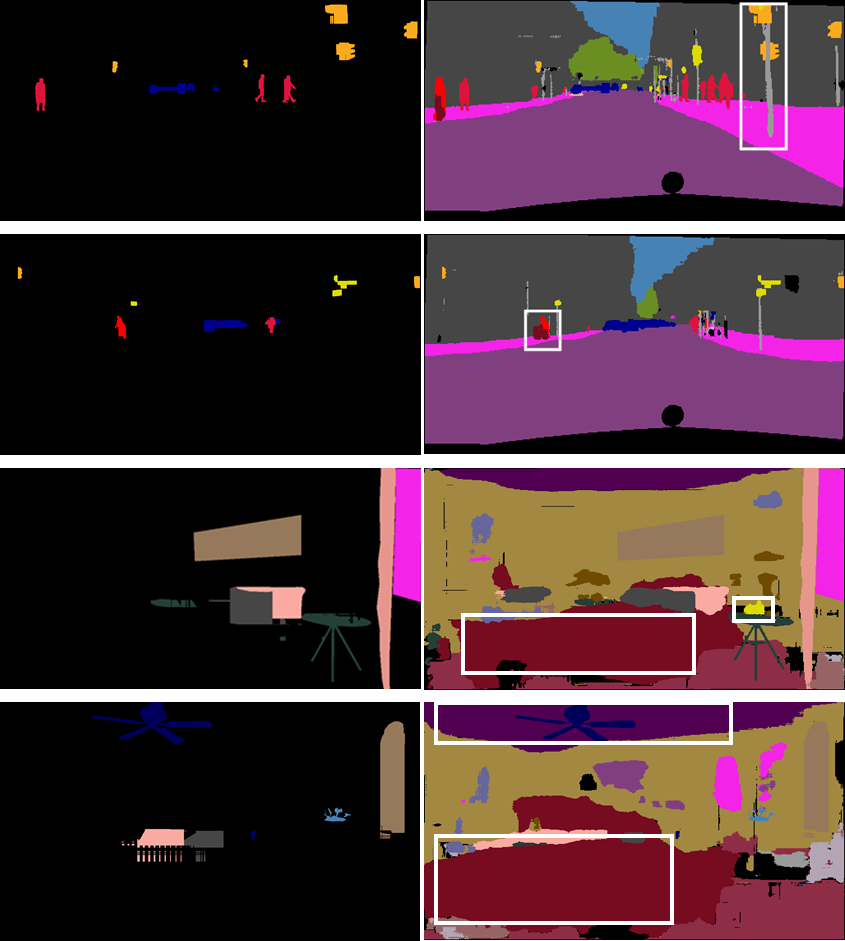}
    \caption{Illustration of our method's ability to capture object co-occurrence relationships.
    Row~1: Poles are produced next to traffic-sign (yellow) and traffic-light (orange); 
    Row~2: Bikes are produced adjacent to bikers;
    Row~3, 4:bed (red) is produced adjacent to pillow (pink); ceiling is produced next to fan (blue) and top of bed. 
    Row~4: clock (yellow) is generated on top of table (green).
    }
    \label{fig:obj_co}
\end{figure}

\begin{figure*}[t]
    \centering
    \includegraphics[width=0.86\linewidth]{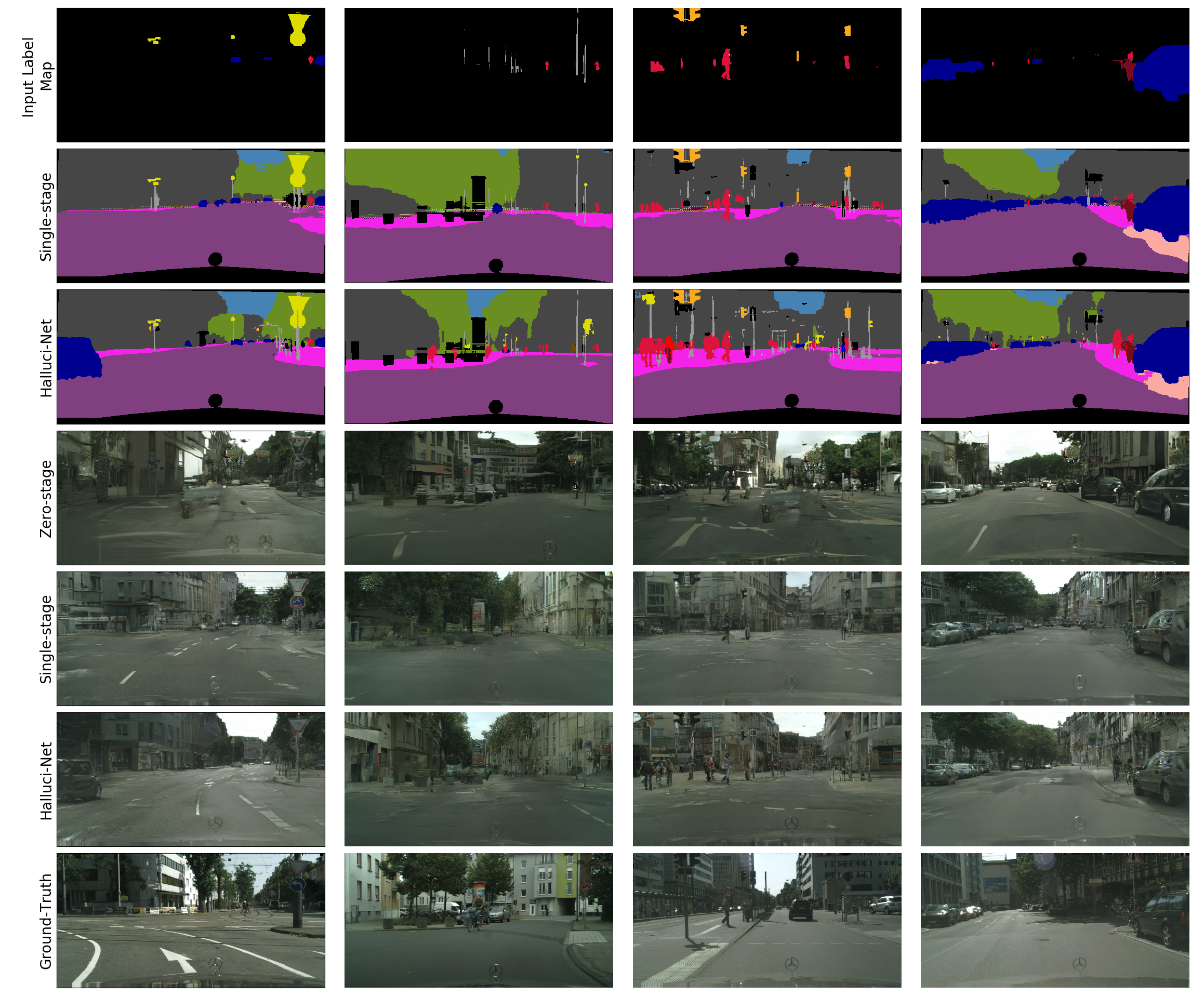}
    
    \caption{Comparison of dense labelmap outputs produced by the two methods.
    As can be seen, the two-stage method produces more details and smoother object shapes.
    The single-stage method is also prone to producing artifacts. 
    Once the dense labelmap is produced, the corresponding image is generated using pix2pixHD. 
    For comparison, we show the ground truth image in the last row. }
    \label{fig:Cityscapes_visual}
\end{figure*}

\begin{figure*}[hbt!]
    \centering
    \includegraphics[ width=0.8\linewidth]{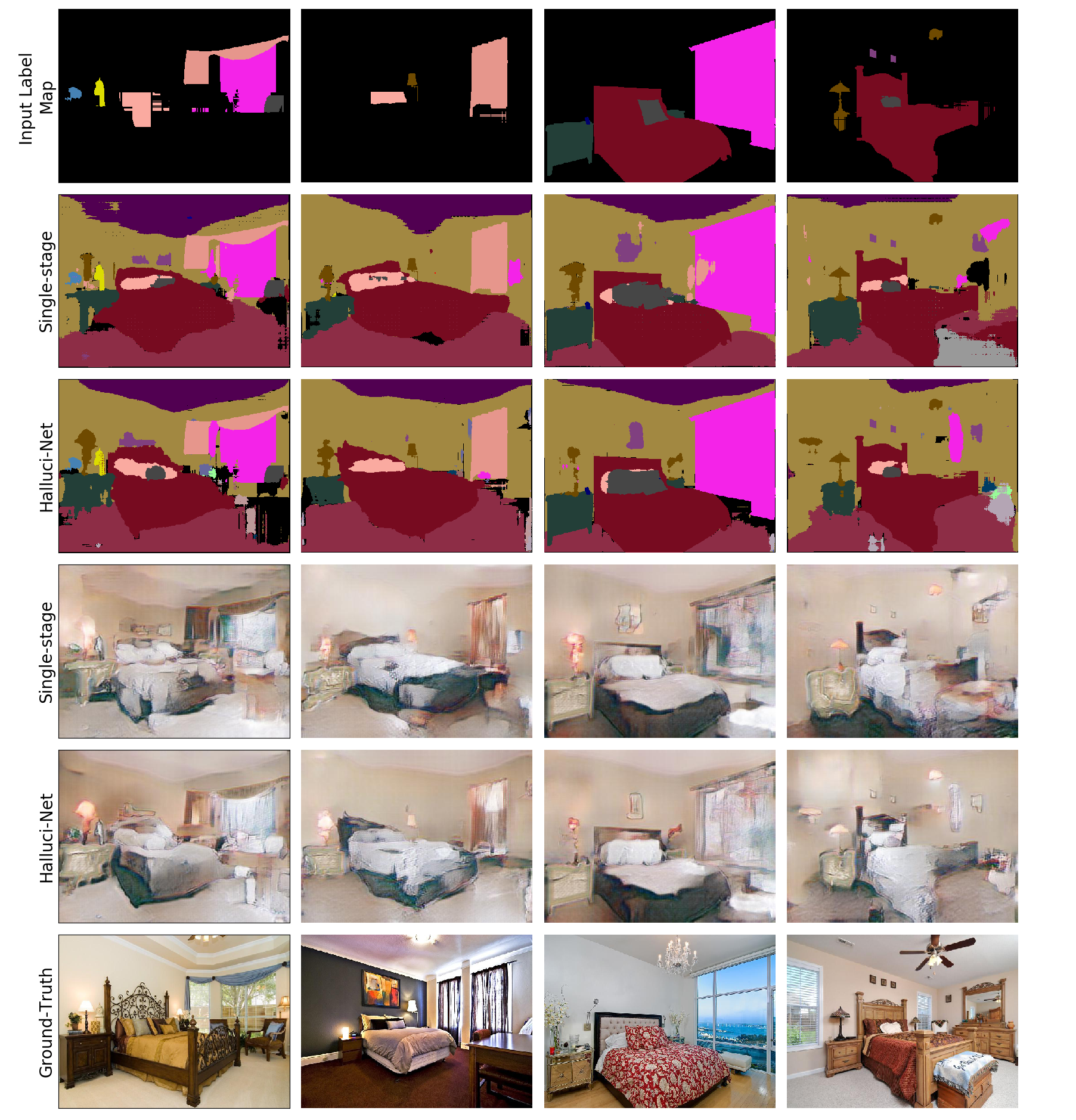}
    \caption{The figure shows the comparison of dense labelmap outputs from the single-stage baseline method and the proposed two-stage method on ADE20K dataset~\cite{zhou2017scene}. As can be seen, the two-stage method produces plausible scene layouts.
    }
    \label{fig:ade20k_visual}
\end{figure*}

\subsection{Quantitative performance}
We evaluate performance of our methods using four metrics, namely Fr\'echet Inception distance (FID), similarity in object co-occurrence statistics, semantic segmentation and a human evaluation study.

\paragraph{Fr\'echet Inception Distance}(FID)\cite{heusel2017gans} is a standard metric used to quantify the fidelity of images generated by GANs. 
FID measures the distance between the distribution of the generated images and the real images in a feature space. 
A pre-trained inception network is used to extract the features corresponding to each image and the mean and covariance of real and generated features $m_r, C_r, m_g, C_g$ respectively are computed.
Then the FID score is calculated as:
\begin{equation}
    FID = ||m_r - m_g||^{2}_{2} + Tr( C_r + C_g -2(C_r C_g)^{\frac{1}{2}})
    \label{fid}
\end{equation}
Table~\ref{table:fid_vs_fsd} shows the FID scores for Halluci-Net for the Cityscapes dataset, in comparison with the zero-stage and single-stage baselines with and without instance boundary, with lower FID score implying superior performance.
Among all approaches, Halluci-Net with instance-boundaries yields the lowest FID score of $40.67$.
Halluci-Net without instance-boundaries also performs considerably better than the single stage approaches. 
While the direct method yields a FID score of $46.21$, it gets progressively better for the single-stage and the proposed Halluci-Net method. 
On ADE20K dataset, Halluci-Net with instance-boundaries has the best FID score of $174.89$ while the single-stage method yields a score of $189.31$.

\noindent{$ $}

\noindent\textbf{Similarity in object co-occurrence:}
In order to evaluate the ability of our method to incorporate object co-occurence in the output, we use the similarity 
measure proposed by Li et. al~\cite{li2019grains}. 
Let $N(c_1)$ be the number of examples in the input dataset in which there is at least one instance of class $c_1$. 
For the corresponding outputs, let $N(c_1,c_2)$ be the number of examples for which there is at least one instance of class $c_2$, that was not present in the input, 
Then we can compute the probability of co-occurence for $c_2$, given $c_1$ and the similarity in co-occurence as:
\begin{align}
    P_{oc}(c_2 | c_1) &= \frac{N(c_1,c_2)}{N(c_1)} \\
    {sim}_{oc}(c_1 | c_2) &= 1 - |P_{oc}^{train}(c_1 | c_2) - P_{oc}^{gen}(c_1 | c_2)|.
\end{align}
Table~\ref{table:obj_co} shows the similarities in co-occurences for different pairs of input and output target classes for Cityscapes.
A score closer to 1 implies high similarity in object co-occurrences between generated and training dataset.

Figure~\ref{fig:obj_co} shows a qualitative analysis of object occurrences captured by our approach in the final output, for Cityscapes as well as ADE20K datasets.
In columns~1 and~2, given that there was a biker in the input, our method is able to appropriately guess the presence and the location of the bike (shown inside the white box). 
Similarly traffic lights are produced at appropriate locations and shape conditioned on the locations of the poles in the input. 
In the bedroom scenes in columns~3 and~4, beds are produced under pillows, ceiling near the fan, and a clock on top of a table.
This evidence suggests that our method is able to capture the inherent object co-occurrence relationships.

\begin{table*}
    \centering
    \small
    \resizebox{\linewidth}{!}{
    \begin{tabular}{@{}lccccccccccc@{}}
        \toprule
        Class & road & side-walk & building & pole & traffic-light & traffic-sign & motorcycle & train & bus & wall & fence\\
        \midrule
        Single-Stage+PSPNet  & 0.9600 & 0.6955 &  0.8472 & 0.4418 & 0.4911 & 0.5913 & 0.3228 & 0.2496 & 0.3845 & 0.0811 & 0.1272\\
        Halluci-Net+PSPNet  &  0.9657 & 0.7400 & 0.8534 & 0.4719 & 0.4906 & 0.5947 & 0.3228 & 0.2249 & 0.4050 & 0.1309 & 0.2948\\  
        \midrule
        Oracle PSPNet    & 0.9801 & 0.8414 &  0.9222 & 0.6399 & 0.7078 & 0.7826 & 0.7764 & 0.7219 & 0.8815 & 0.5208 & 0.5968\\
        \bottomrule
    \end{tabular}
    }
    \caption{Segmentation scores (mIoU)
    on Cityscapes dataset~\cite{cordts2016Cityscapes} for different object categories.
    Halluci-Net performs better than the single stage approach in most cases.
    }
    \label{table:segment_score_hall_obj}
\end{table*}
\begin{table}
    \centering 
    \small
    \begin{tabular}{lccc}
    \toprule
    Method & mIoU & mAcc & Accuracy \\
    \midrule
    Single-Stage+PSPNet  & 0.5111 & 0.6215 &  0.9131 \\
    Halluci-Net+PSPNet  &  0.5501 & 0.6647 & 0.9214 \\ 
    \midrule
    Oracle PSPNet    &  0.7730 & 0.8431 &  0.9597\\
    \bottomrule
    \end{tabular}
    \caption{Segmentation scores in terms of 
    mean Intersection over Union (mIoU) and per-pixel accuracies on Cityscapes dataset~\cite{cordts2016Cityscapes} with PSPNet trained on the generated label-image pairs from different methods. 
    The scores are obtained on the label-image pairs from the original validation set. 
    }
    \label{table:segment_score_hall}
\end{table}

\noindent{$ $}

\noindent\textbf{Semantic Segmentation Score:}
To evaluate the utility and quality of the generated labelmaps and generated images, we train a semantic segmentation network~\cite{zhao2017pyramid} (with ResNet50 as the backbone) using the generated pairs, and test on the original validation set of images. 
Halluci-Net performs better than the Single-Stage baseline in terms of all standard semantic segmentation metrics as shown in Table~\ref{table:segment_score_hall}.
For comparison, we show the results using the Oracle PSPNet, that is trained on the original labelmap--image pairs. 
We also show the mIoU accuracies for different object categories in Table~\ref{table:segment_score_hall_obj}.
It can be seen that the Halluci-Net+PSPNet performs better than Single-Stage+PSPNet for most object categories, except ``train". 
However, for classes like train, motorcycle, wall and fence, the proposed method performs poorly as compared to the Oracle PSPNet. 
This shows the limitation of our method to generate less-frequent object categories. \\

\noindent{\textbf{Human Evaluation Study:}}
In addition to this, we perform a human evaluation study on the generated labelmaps from our method and the baseline method. We show 200 randomly selected pairs of labelmaps to five human workers, and asked them to tell which of the two labelmaps they prefer in terms of shape of the objects that are generated, how much physical sense they make. Across the five human workers, our method was preferred 116 times and the baseline method 84 times.

\begin{figure}[t]
     \centering
     \includegraphics[width=\linewidth]{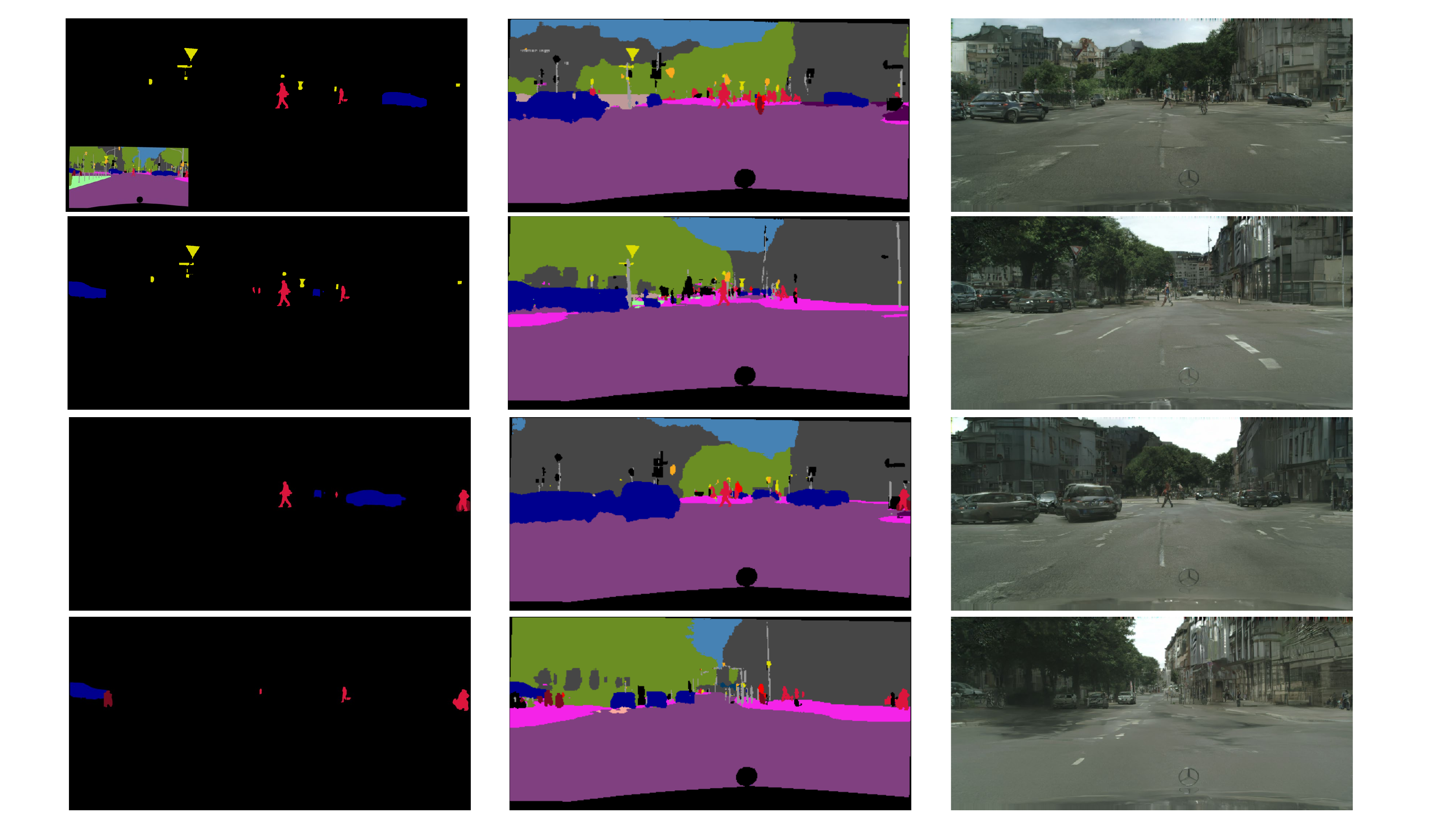}
    \rule[1ex]{\linewidth}{0.1pt}
     \includegraphics[width=\linewidth]{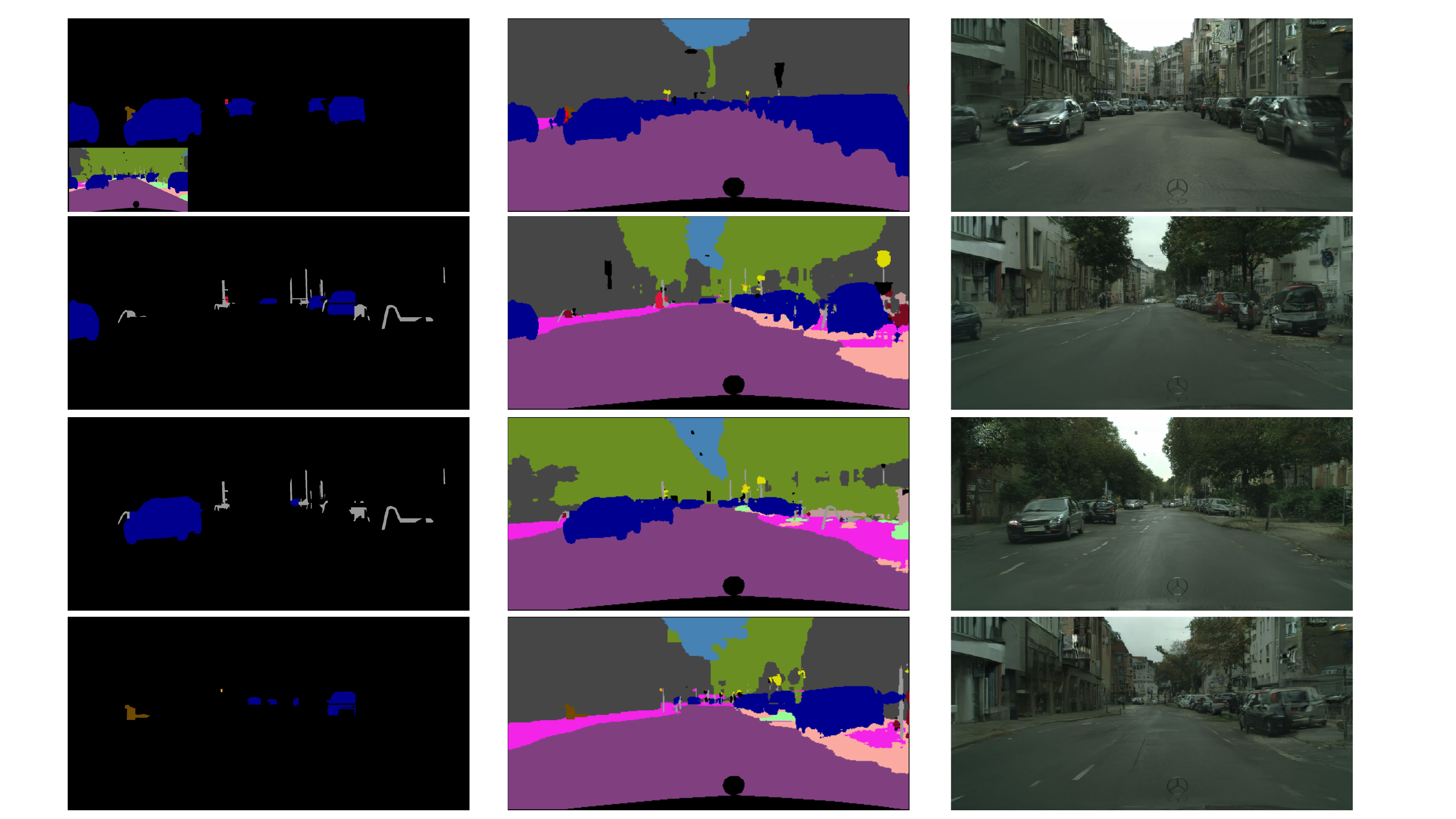}
     \caption{The figure shows how it is possible to generate multiple novel scene layouts, given a parent dense scene layout (inset). \\
     Column~1: four different randomly sampled sparse labelmaps. 
     Column~2: different dense labelmaps produced by the proposed method. 
     Column~3: corresponding synthesized image.}
    \label{fig:mult}
\end{figure}

\subsection{Visual Results}
We show visual results for scene layout synthesis as well as for image synthesis by comparing Halluci-Net with the two baseline methods: zero-stage (direct mapping from the sparse label to image) and single-stage method.
Results for the Cityscapes dataset are shown in Figure~\ref{fig:Cityscapes_visual}. 
Row~1 shows the input labelmap. 
It can be seen from row~2 and row~3 that our method produces scene layouts with more details and smoother and precise object shapes, as compared to the baseline method. 
In rows~4, 5 and 6, we show the images generated for direct method, Single-Stage+pix2pixHD and Halluci-Net+pix2pixHD. 
With sparse inputs, the zero-stage method is prone to produce artifacts in the images that are generated as can be seen in column~1 and~3. As compared to direct method and
Single-Stage+pix2pixHD, Halluci-Net+pix2pixHD produces images with richer content owing to the superior scene layout created by the two-stage network. 
For comparison we have shown the ground-truth expected image. 
Similarly, in Figure~\ref{fig:ade20k_visual}, we show the synthesized scene layouts and the corresponding images for the baseline Single-Stage method and the proposed Halluci-Net method on ADE20K dataset. 
Our method is able to generate plausible scene layouts.

\subsection{Single Layout to Multiple Layouts}
We propose a method to interactively generate several scene layouts from a single dense layout. In Figure~\ref{fig:mult} we show how a user can edit a dense scene layout. The user can choose to sample different object instances. Column~1 shows four such random sampling of object instances from the parent scene layout. Column~2 shows the different scene layouts that were generated using Halluci-Net and finally images generated are shown in Column~3. 
This simple experiment shows the potential of our method to generate novel scene layouts.
multiple variants for the complete scene can be synthesized.
These algorithmically generated images can be used by content creators directly, or indirectly as templates to guide paintings or drawings on conventional art media.

\section{Conclusion}
We introduced the new problem of generated complex scenes, composed of several objects from extremely sparse labelmaps and proposed a two-stage network approach to generate the dense labelmap. 
We show both qualitatively and quantitatively that the proposed method is able to produce high-quality dense labelmaps for street traffic images in Cityscapes dataset and bed-room images in ADE20K dataset. 
While the results are highly encouraging, we also note that our method is limited in its ability to generate objects of rare classes like wall, fence, as demonstrated earlier in terms of low segmentation accuracy. 

\section*{Broader Impact}
Image synthesis with constraints has broad applications in consumer and industrial applications such as autonomous driving.
Application areas in the consumer media realm include image-synthesis for graphics, arts, and social-media.
Further, image synthesis is also being currently used to collect training-sets for domains in which data collection may be difficult.
In this situation, there is the open question of whether machine-learning techniques trained on synthetic imagery are trustworthy in critical deployments. 
Image synthesis approaches that can generate very realistic looking pictures can also have unintended effects such as in the spread of disinformation. 
To mitigate these unintended effects, there is a parallel body of technical work in detecting sources of disinformation in synthesized imagery.  
Further, there is an implicit reliance on large computational resources, which in turn implies financial resources. 
Additional avenues of research to investigate the ethical and legal implications of using image-synthesis techniques, and whether they lead to trustworthy systems, should be encouraged with use-inspired interdisciplinary teams.

\section*{Acknowledgments}
This work was supported by a gift from Adobe Inc.\ to the Geometric Media Lab.  
The work of KK, TG, and AS was supported by ARO Grant W911NF-16-1-0441.

{\small
\bibliographystyle{ieee_fullname}
\bibliography{egbib}
}

\end{document}